\documentclass[a4paper]{spie}  %>>> use this instead for A4 paper  
\usepackage[pdftex]{graphicx,xcolor}

\usepackage{amsmath}
\usepackage{cite}
\usepackage{booktabs}
\usepackage{multirow}
\usepackage{tabularx}
\usepackage{url}
\usepackage{adjustbox}
\usepackage{array}
\usepackage{overpic}
\usepackage{colortbl}
\usepackage[colorlinks=true, allcolors=blue]{hyperref}
\usepackage{ulem}
\usepackage{setspace}
\usepackage{float} % stop figure from floating with [H] (package {here} is obsolete
\usepackage{subfig}
\usepackage{hyperref}
\title{Towards dense volumetric pancreas segmentation in CT using 3D fully convolutional networks}
\author[a]{Holger Roth}
\author[a]{Masahiro Oda}
\author[b]{Natsuki Shimizu}
\author[b]{Hirohisa Oda}
\author[a]{Yuichiro Hayashi}
\author[c]{Takayuki Kitasaka}
\author[d]{Michitaka Fujiwara}
\author[e]{Kazunari Misawa}
\author[f,b]{Kensaku Mori}

\affil[a]{Graduate School of Informatics, Nagoya University, Japan}
\affil[b]{Graduate School of Information Science, Nagoya University, Japan}
\affil[c]{Faculty of Information Science, Aichi Institute of Technology, Japan}
\affil[d]{Nagoya University Graduate School of Medicine, Japan}
\affil[e]{Department of Gastroenterological Surgery, Aichi Cancer Center Hospital, Japan}
\affil[f]{Information \& Communications, Nagoya University, Japan}

\authorinfo{Send correspondence to: \href{mailto:rothhr@mori.m.is.nagoya-u.ac.jp}{rothhr@mori.m.is.nagoya-u.ac.jp} and \href{mailto:kensaku@is.nagoya-u.ac.jp}{kensaku@is.nagoya-u.ac.jp}}

\begin{document}
\maketitle
%%%%%%%%%%%%%%%%%%%%%%%%%%%%%%%%%%%%%%%%%%%%%%%%%%%%%%%%%%%%%%%%%%%%%%%%%%%%%%%%%%%%%%%%%%%%%%%%
%%%%%%%%%%%%%%%%%%%%%%%%%%%%%%%%%%%%%%%%%%%%%%%%%%%%%%%%%%%%%%%%%%%%%%%%%%%%%%%%%%%%%%%%%%%%%%%%
\begin{abstract}
Pancreas segmentation in computed tomography imaging has been historically difficult for automated methods because of the large shape and size variations between patients. In this work, we describe a custom-build 3D fully convolutional network (FCN) that can process a 3D image including the whole pancreas and produce an automatic segmentation. We investigate two variations of the 3D FCN architecture; one with concatenation and one with summation skip connections to the decoder part of the network. We evaluate our methods on a dataset from a clinical trial with gastric cancer patients, including 147 contrast enhanced abdominal CT scans acquired in the portal venous phase. Using the summation architecture, we achieve an average Dice score of 89.7 $\pm$ 3.8 (range [79.8, 94.8])\% in testing, achieving the new state-of-the-art performance in pancreas segmentation on this dataset.
\end{abstract}


%%%%%%%%%%%%%%%%%%%%%%%%%%%%%%%%%%%%%%%%%%%%%%%%%%%%%%%%%%%%%%%%%%%%%%%%%%%%%%%%%%%%%%%%%%%%%%%%
%%%%%%%%%%%%%%%%%%%%%%%%%%%%%%%%%%%%%%%%%%%%%%%%%%%%%%%%%%%%%%%%%%%%%%%%%%%%%%%%%%%%%%%%%%%%%%%%
\section{INTRODUCTION}
Automated segmentation of the pancreas in computed tomography (CT) imaging is challenging because of the larger shape and size variations between patients compared to other organs, and because of the low contrast to surrounding tissues \cite{roth2017spatial}. Recent advantages have mainly been due to the application of deep learning based methods that allow the efficient learning of features from the imaging data directly. Especially, the development of fully convolutional neural networks (FCN) \cite{long2015fully} has further raised the state-of-the-art in semantic segmentation significantly. However, current deep learning methods for pancreas segmentation were still limited to either 2D image processing \cite{roth2017spatial,zhou2016pancreas} or subvolume processing \cite{roth2017hierarchical} which can introduce discontinuity artifacts, mainly due to the memory restrictions caused by moving to full 3D volume processing. In this work, we describe a 3D FCN that can process a 3D image including the whole pancreas and segment it with the highest accuracy reported in literature to date, making advantage of state-of-the-art FCN architectures and multi-GPU processing.
\section{METHODS}
%%%%%%%%%%%%%%%%%%%%%%%%%%%%%%%%%%%%%%%%%%%%%%%%%%%%%%%%%%%%%%%%%%%%%%%%%%%%%%%%%%%%%%%%%%%%%%%%
\begin{figure}[tb]
  \centering
  \subfloat{\adjincludegraphics[valign=c,width=0.3\textwidth]{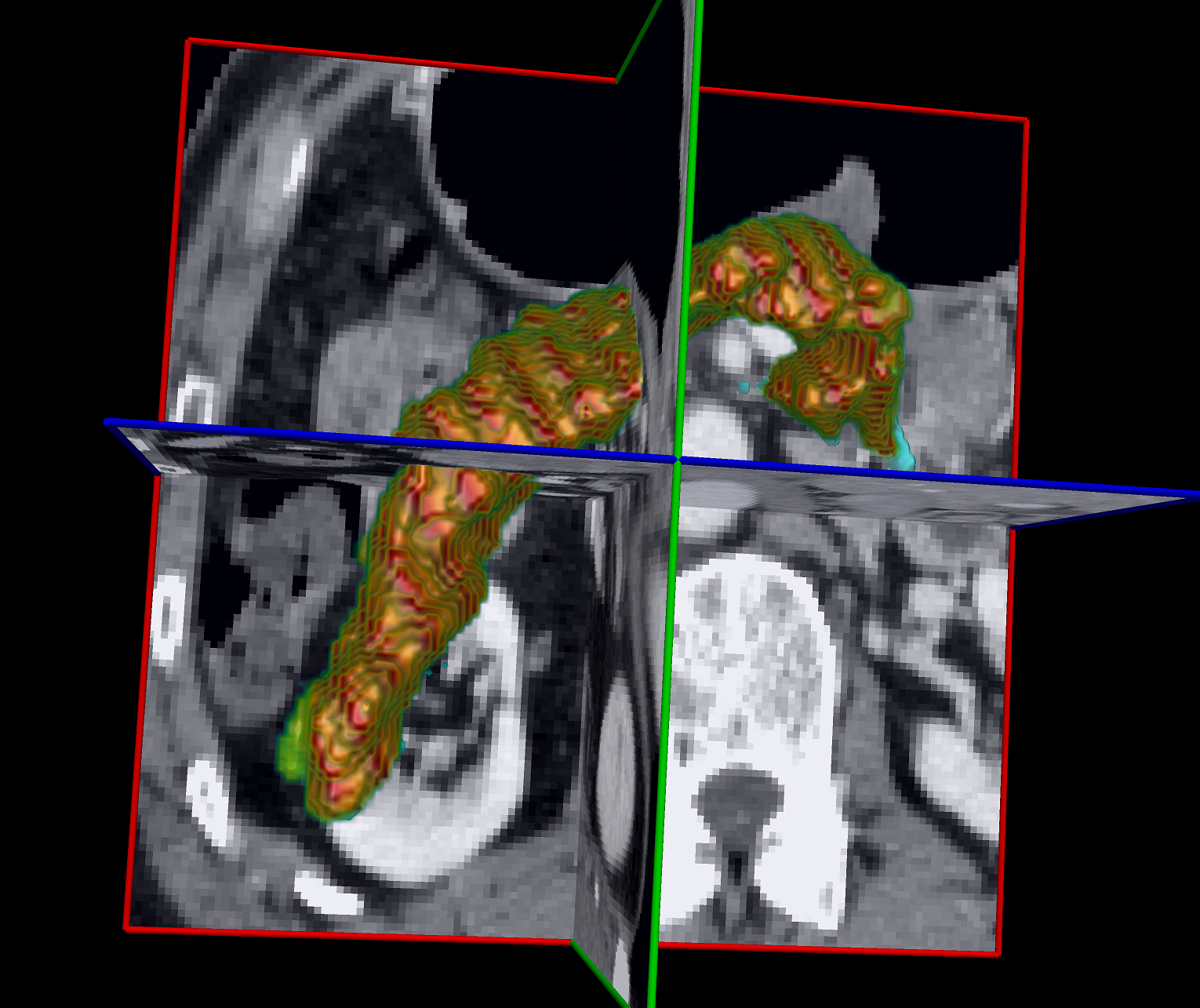}}
  %\hfill
  \subfloat{\adjincludegraphics[valign=c,width=0.3\textwidth]{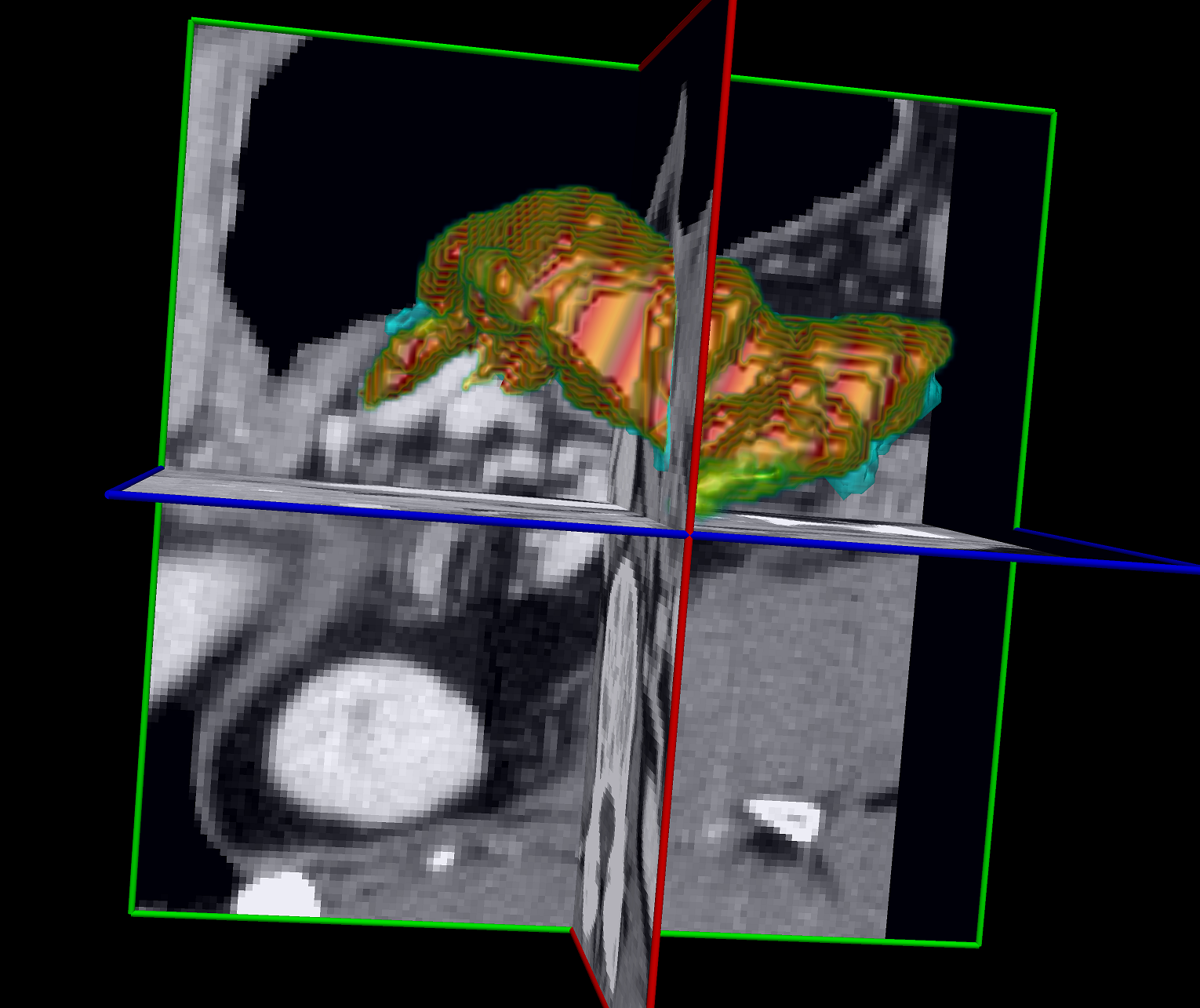}}    
  %\hfill
  \subfloat{\adjincludegraphics[valign=c,width=0.3\textwidth]{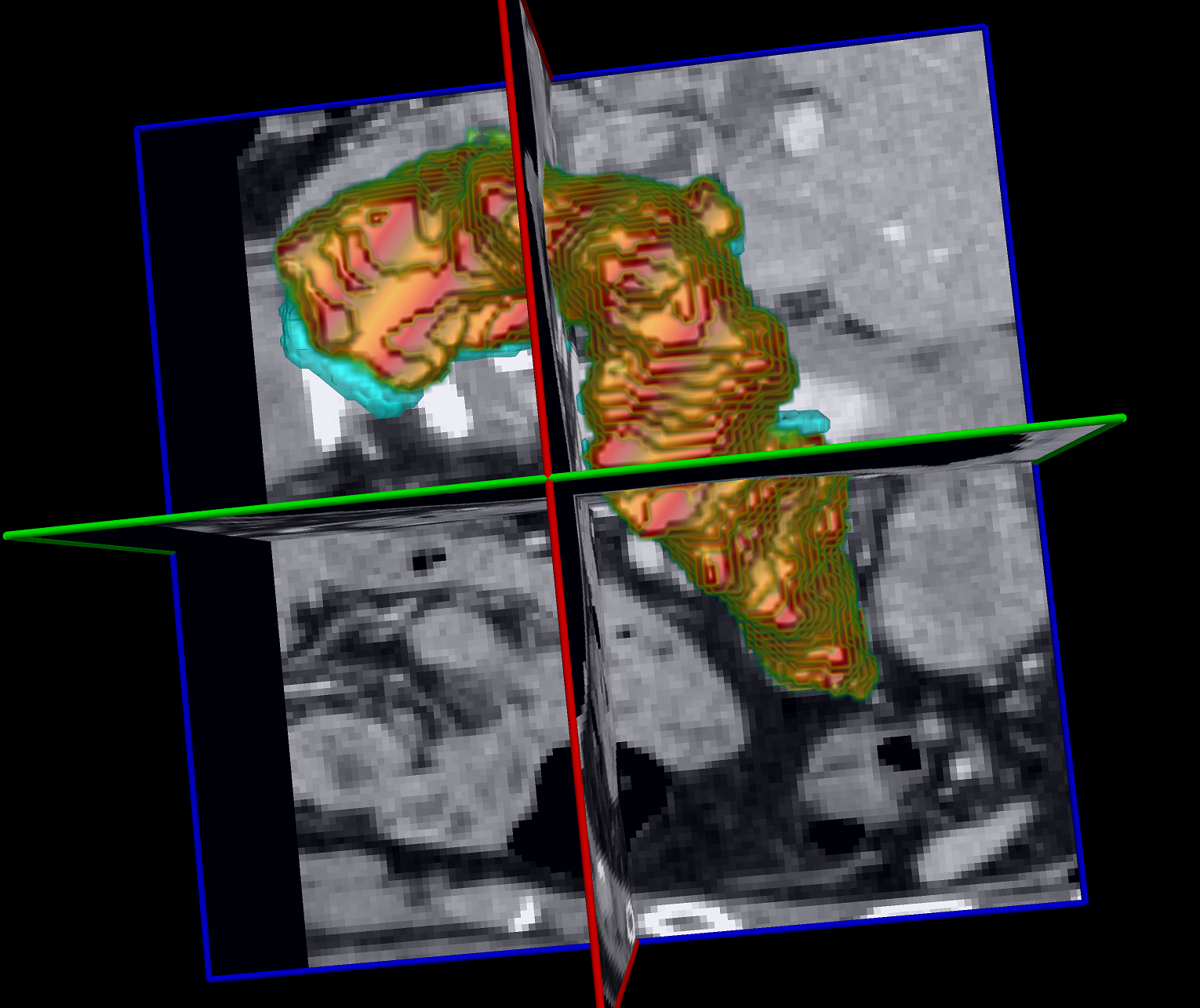}}  
  %\hfill
  \subfloat{\adjincludegraphics[valign=c,width=0.051\textwidth]{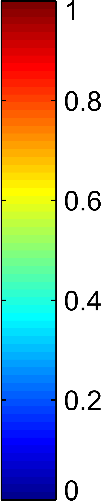}} 
  \vspace{1em}
\caption{ 3D visualization of the resulting pancreas segmentation in CT from three views. The gold standard segmentation of the pancreas is shown in cyan. Our predicted probability score for each voxel in the images is rendered in jet color scale. \label{fig:result_3D_view}}
\end{figure}
%%%%%%%%%%%%%%%%%%%%%%%%%%%%%%%%%%%%%%%%%%%%%%%%%%%%%%%%%%%%%%%%%%%%%%%%%%%%%%%%%%%%%%%%%%%%%%%%
Recent advances in fully convolutional networks (FCN) have made it feasible to train models for pixel-wise semantic segmentation in an end-to-end fashion \cite{long2015fully}. Efficient implementations of 3D convolution and growing GPU memory have made it possible to extent these methods to 3D medical imaging and train networks on large amounts of annotated volumes. One such example is the recently proposed 3D U-Net \cite{cciccek20163d}, which applies a 3D FCN with skip connections to sparsely annotated biomedical images. In the following, we extent the 3D U-Net architecture to our problem and utilize a random forest based bounding box locator in order to simplify the problem of initial pancreas localization. The main contribution is the dense segmentation of the pancreas using a direct mapping between the input volume and the segmentation via 3D FCNs. Example results of our method can be seen in Fig. \ref{fig:result_3D_view}. 

\subsection{Candidate region generation} We utilize a state-of-the-art random forest (RF) regression method to give us an initial bounding box region of the pancreas, while removing vast amounts of the background of the image. In order to guarantee 100\% recall, we add a margin to the automatically generated bounding boxes. Our RF repressor was developed in previous work \cite{oda2016regression} and allows robust localization of the pancreas in our dataset. Bounding box detection helps reducing the input candidate space that our FCN needs to process in order to efficiently learn to differentiate pancreas from the surrounding anatomy.

\subsection{Fully convolutional networks (3D U-Net)} FCNs have the ability to solve challenging classification tasks in a data-driven manner, given a training set of images and labels $S={(I_n,L_n ),n=1,\dots,N}$, where $I_n$ denotes the raw CT images and $L_n$ denotes the ground truth label images. This allows the network to find a direct mapping from the image to segmentation space by learning a very complex non-linear function between the two. A popular recent FCN architecture is 3D U-Net \cite{cciccek20163d}, which utilizes transposed convolutions to remap the lower resolution feature maps within the network to the denser space of the input images. The 3D U-Net architecture consists of symmetric analysis and synthesis paths with four resolution levels each. Each resolution level in the analysis path contains two $3\times3\times3$ convolutional layers, each followed by rectified linear units (ReLu) and a $2\times2\times2$ max pooling with strides of two in each dimension. In the synthesis path, the convolutional layers are replaced by up-convolutions\footnote{We realize the up-convolutional layers as up-sampling layer followed by a convolutional layer.} of $2\times2\times2$ with strides of two in each dimension. These are followed by two $3\times3\times3$ convolutions, each of which has a ReLu. Furthermore, 3D U-Net employs shortcut (or skip) connections from layers of equal resolution in the analysis path to provide higher-resolution features to the synthesis path \cite{cciccek20163d}. The final convolutional layer employs a sigmoid activation function to compute a probability map for pancreas as output of the network.

\subsection{Extended 3D U-Net} 
\begin{figure}[tb]
  \centering
  \adjincludegraphics[width=0.95\textwidth]{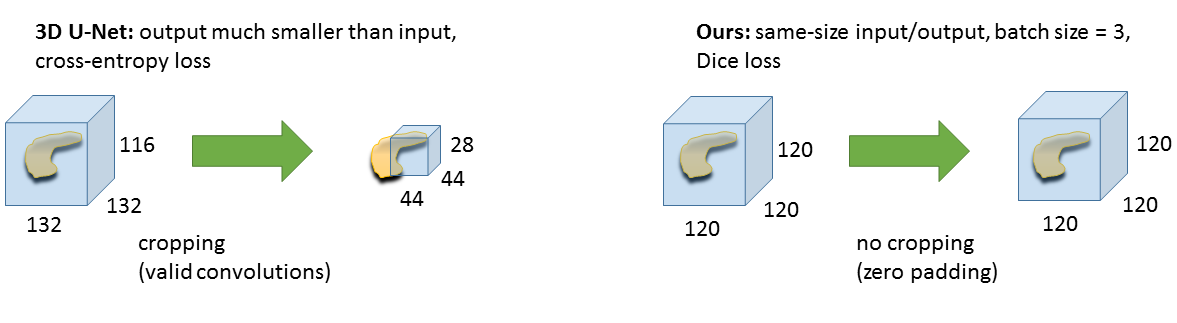}
\caption{ The proposed architecture of our dense network (right) in comparison to 3D U-Net\cite{cciccek20163d} (left) with a pancreas sketch for illustration. 3D U-Net has been designed to run efficiently on one 12 GB GPU. It only utilizes valid convolutions and does have to use cropping in order to fit the network into memory, which can cause discontinuities in the segmentation result \cite{roth2017hierarchical}. On the other hand, multi-GPU processing allows us to employ a full end-to-end architecture that projects that segments the full 3D volume directly using same size convolution (via zero padding). This allows us to cover the full pancreas in the predicted subvolume. \label{fig:network_proposal}}
\end{figure}

In this work, we develop a custom-build 3D U-Net type FCN that can be employed on multiple GPUs, allowing mini-batch training, and use of a larger amount of GPU memory. Due to memory constraints, when applying 3D U-Net on one GPU card, the output size of the network had to be significantly smaller than the input size \cite{cciccek20163d}. By utilizing multiple GPUs (four in our case), we can define a network that allows a full end-to-end segmentation of the 3D volume covering the whole pancreas, as illustrated in Fig. \ref{fig:network_proposal}. The next advantage of using multiple GPUs is that we can now fit volume batch sizes $>$1 into memory at each training iteration in order to use mini-batch updates and batch normalization \cite{ioffe2015batch}. We investigate two variations of the architecture, one with concatenation (used by the original 3D U-Net) and one simplified version using summation skip connections to the encoder part of the network \cite{drozdzal2016importance}. Using concatenation, our network has ~25M parameters; using summation, this reduces to ~12M parameters at an input size of $120\times120\times120$. Our use of summation layers mimics residual learning, which has been shown to improve deep neural network training in practice \cite{he2016deep}.
%%%%%%%%%%%%%%%%%%%%%%%%%%%%%%%%%%%%%%%%%%%%%%%%%%%%%%%%%%%%%%%%%%%%%%%%%%%%%%%%%%%%%%%%%%%%%%%%
%%%%%%%%%%%%%%%%%%%%%%%%%%%%%%%%%%%%%%%%%%%%%%%%%%%%%%%%%%%%%%%%%%%%%%%%%%%%%%%%%%%%%%%%%%%%%%%%
\subsection{Data augmentation} 
\begin{figure}[htb]
  \centering
  \subfloat[]{\adjincludegraphics[width=0.35\textwidth]{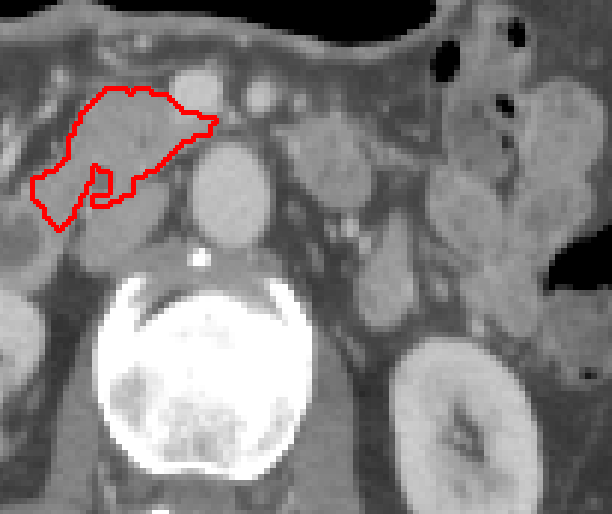}}
  %\hfill
	\hspace{2em}
  \subfloat[]{\adjincludegraphics[width=0.35\textwidth]{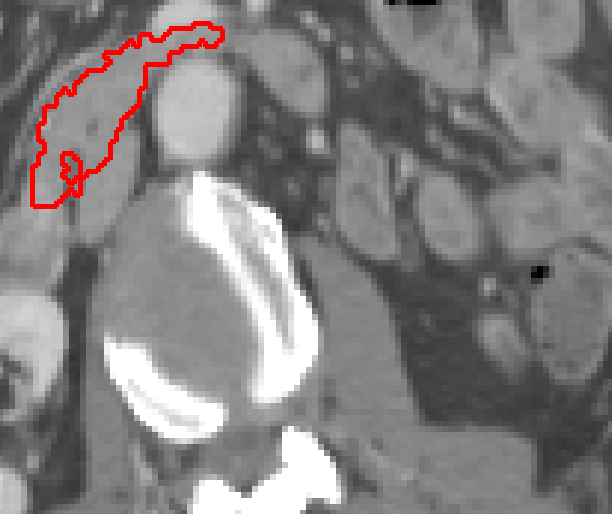}}    
    \vspace{1em}
\caption{ Cross-section through the same patient CT image (a) before and (b) after random deformation with the training ground truth overlaid in red. \label{fig:deform}}
\end{figure}

In training, we employ smooth B-spline deformations to both the image and label data, similar to \cite{cciccek20163d}. The deformation fields are randomly sampled from a uniform distribution with a maximum displacement of 4 and a grid spacing of 24 voxels. Furthermore, we applied random rotations between $-20^{\circ}$ and $+20^{\circ}$, and translations of -20 to 20 voxels in each dimension at each iteration in order to generate plausible deformations during training. Some example deformations are shown in Fig. \ref{fig:deform}. This type of data augmentation can help to artificially increase the training data and encourages convergence to more robust solutions while reducing overfitting to the training data (see Fig. \ref{fig:training}). 

\subsection{Implementation} All models are implemented in Keras with the TensorFlow backend, and use Adam optimization \cite{kingma2014adam} with an initial learning rate of 1e-2. As optimization function, we utilize the pseudo Dice score proposed in \cite{milletari2016v}. We train all models for 20,000 iterations, which takes about one week. Inference, however, is achieved in less than 10 seconds per case. A DeepLearning BOX (uniV) including four NVIDIA Quadro P6000 GPUs with 24 GB each is used for the multi-GPU experiments.

%%%%%%%%%%%%%%%%%%%%%%%%%%%%%%%%%%%%%%%%%%%%%%%%%%%%%%%%%%%%%%%%%%%%%%%%%%%%%%%%%%%%%%%%%%%%%%%%
%%%%%%%%%%%%%%%%%%%%%%%%%%%%%%%%%%%%%%%%%%%%%%%%%%%%%%%%%%%%%%%%%%%%%%%%%%%%%%%%%%%%%%%%%%%%%%%%
%%%%%%%%%%%%%%%%%%%%%%%%%%%%%%%%%%%%%%%%%%%%%%%%%%%%%%%%%%%%%%%%%%%%%%%%%%%%%%%%%%%%%%%%%%%%%%%%
\section{EXPERIMENTS \& RESULTS}
\subsection{Data}
Our dataset originates from a research study with gastric cancer patients. 147 contrast enhanced abdominal CT scans were acquired in the portal venous phase. Each CT volume consists of 263−1061 slices of $512\times512$ pixels. Voxel dimensions are [0.55-0.82, 0.55-0.82, 0.4-0.80] mm. The pancreas was delineated by three trained researchers and confirmed by a clinician. After RF regression, the detected bounding box regions are resampled to a fixed size of $120\times120\times120$ and fed to our 3D FCN models. 
\subsection{Evaluation}
\begin{figure}[tb]
	\centering
	\adjincludegraphics[width=0.95\textwidth]{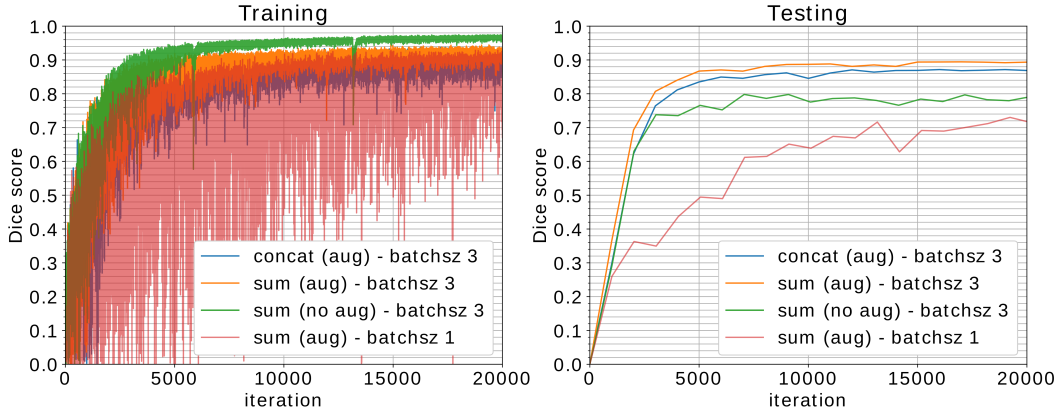}
	\caption{Comparison between use of concatenation or summation skip connections (with and without data augmentation) during training and testing. We plot the average Dice score of the models at each training iteration. An advantage of using summation skip connections and larger batch sizes can be observed in testing. Without data augmentation (\textbf{no aug}), the model overfits to the training data. \label{fig:training}}
\end{figure}

\begin{figure}[tb]
	\centering
	\subfloat[Ground truth]{\adjincludegraphics[width=0.245\textwidth]{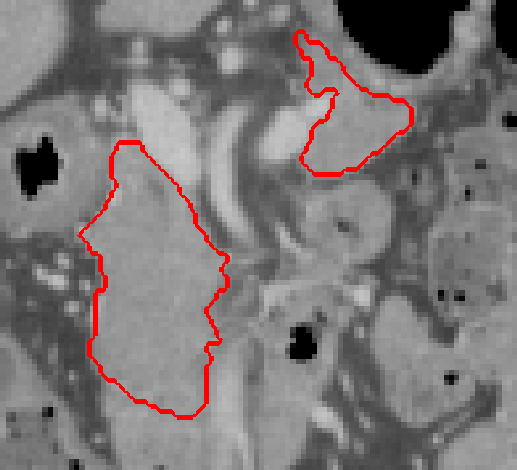}}
	\hfill
	\subfloat[sum (no aug)]{\adjincludegraphics[width=0.245\textwidth]{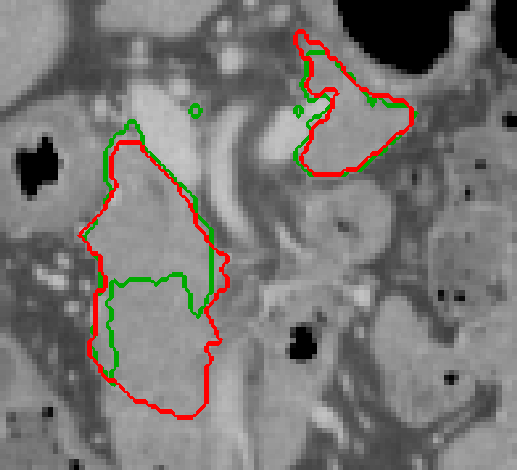}}     
	\hfill  			
	\subfloat[concat (aug)]{\adjincludegraphics[width=0.245\textwidth]{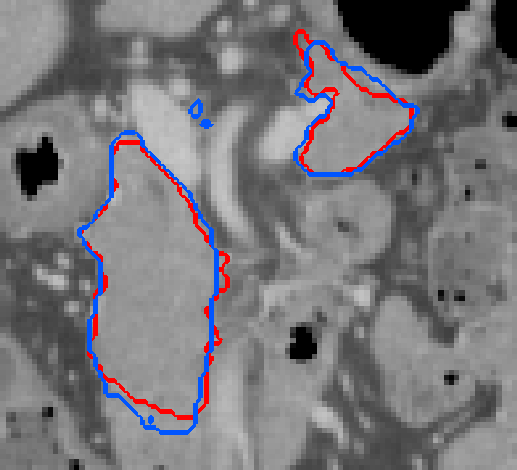}} 	
	\hfill
	\subfloat[sum (aug)]{\adjincludegraphics[width=0.245\textwidth]{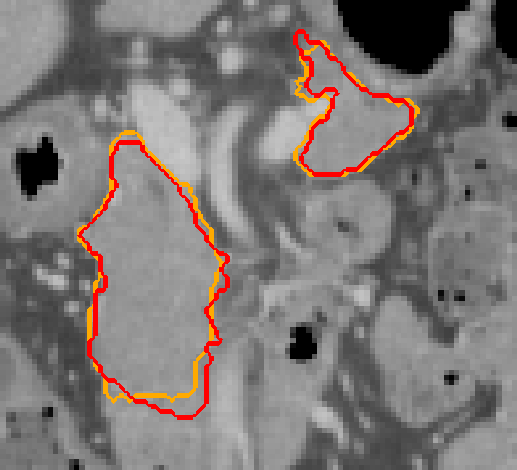}}    	
	\vspace{1em}
	
	\subfloat[Ground truth]{\adjincludegraphics[width=0.245\textwidth]{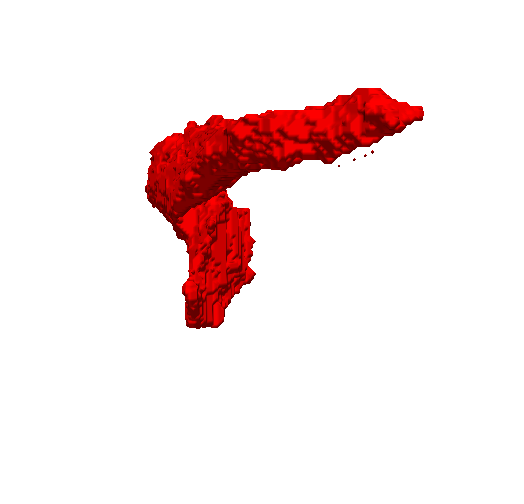}}
	\hfill
	\subfloat[sum (no aug)]{\adjincludegraphics[width=0.245\textwidth]{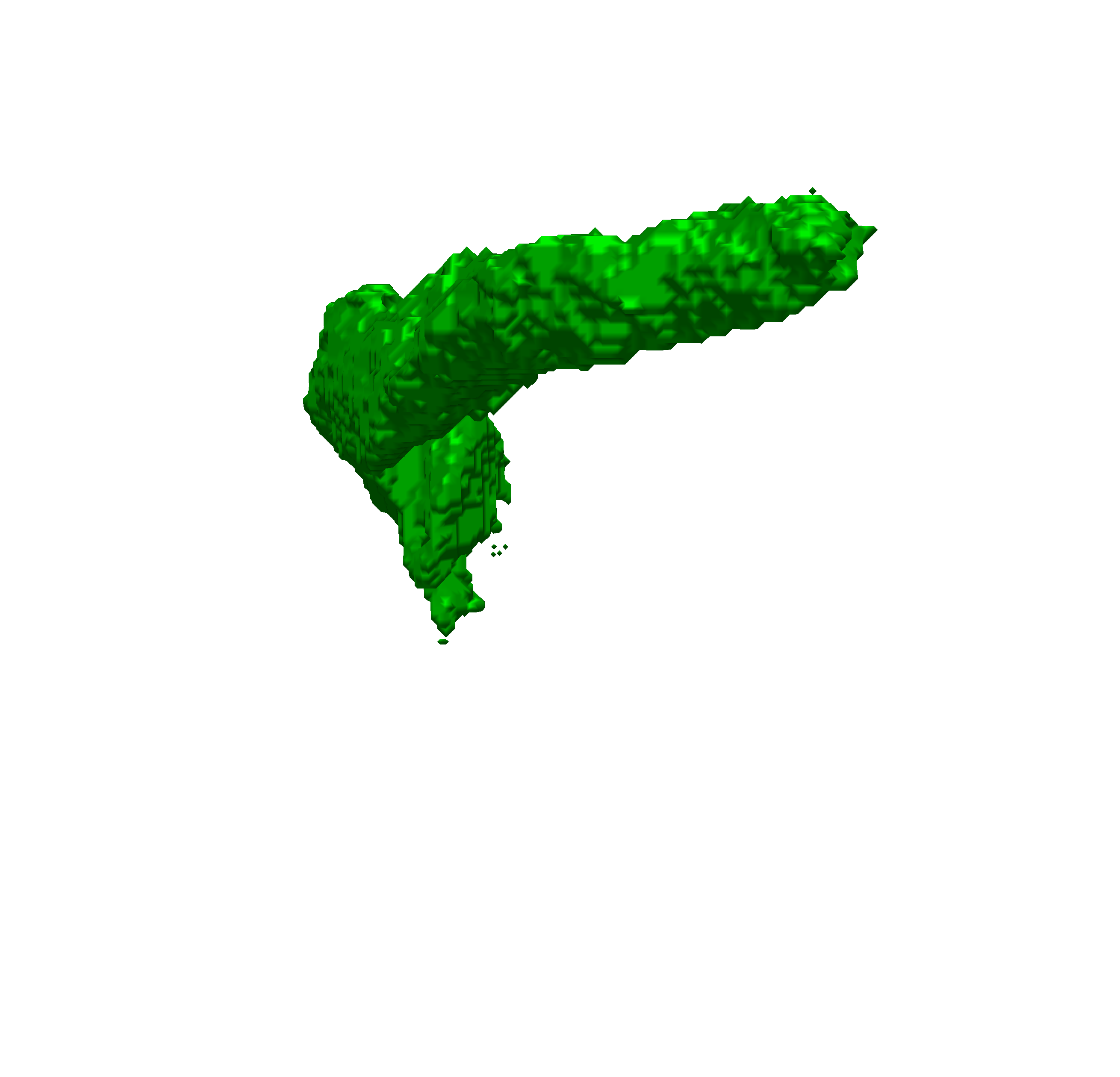}}     
	\hfill  			
	\subfloat[concat (aug)]{\adjincludegraphics[width=0.245\textwidth]{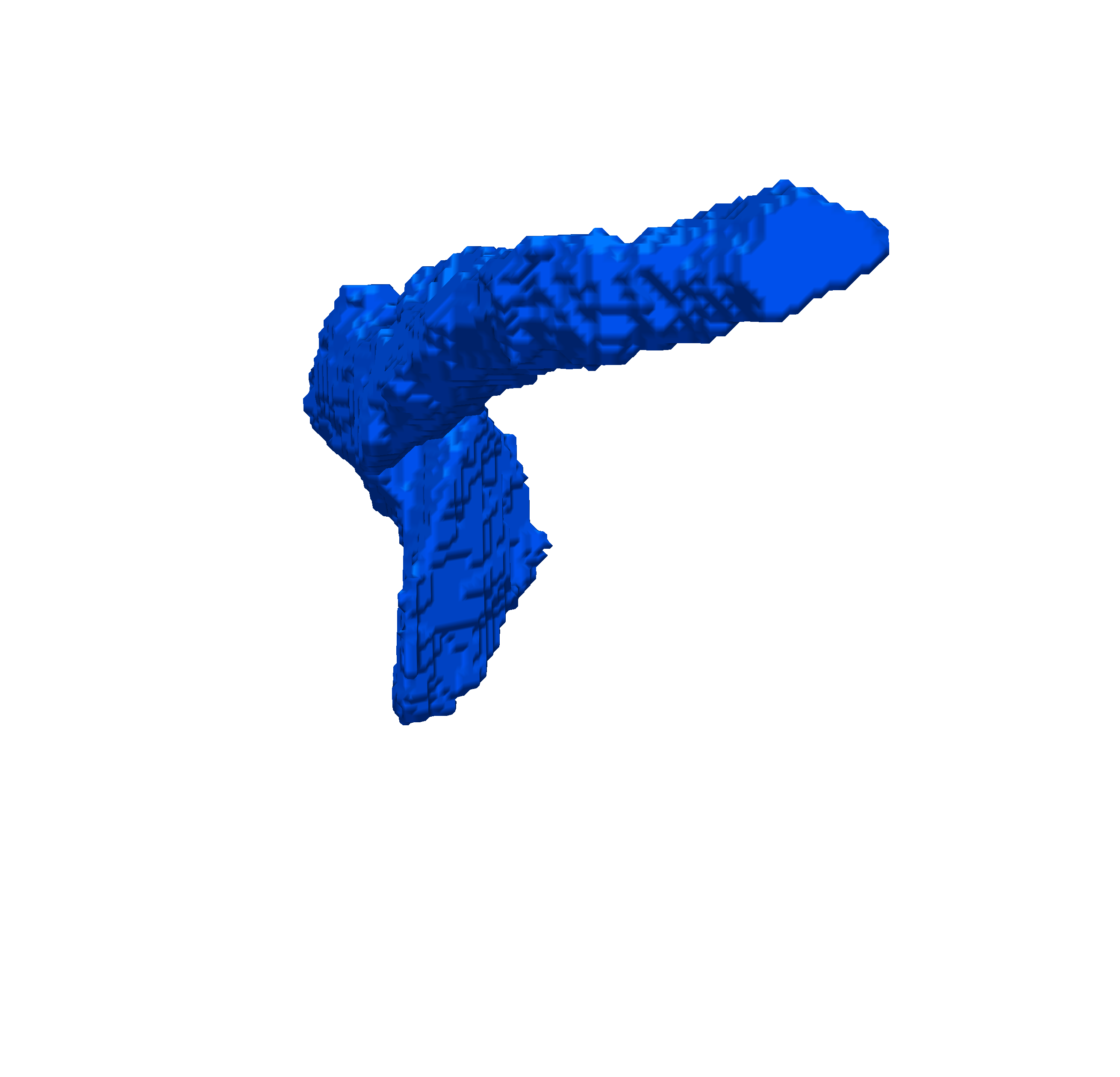}} 	
	\hfill
	\subfloat[sum (aug)]{\adjincludegraphics[width=0.245\textwidth]{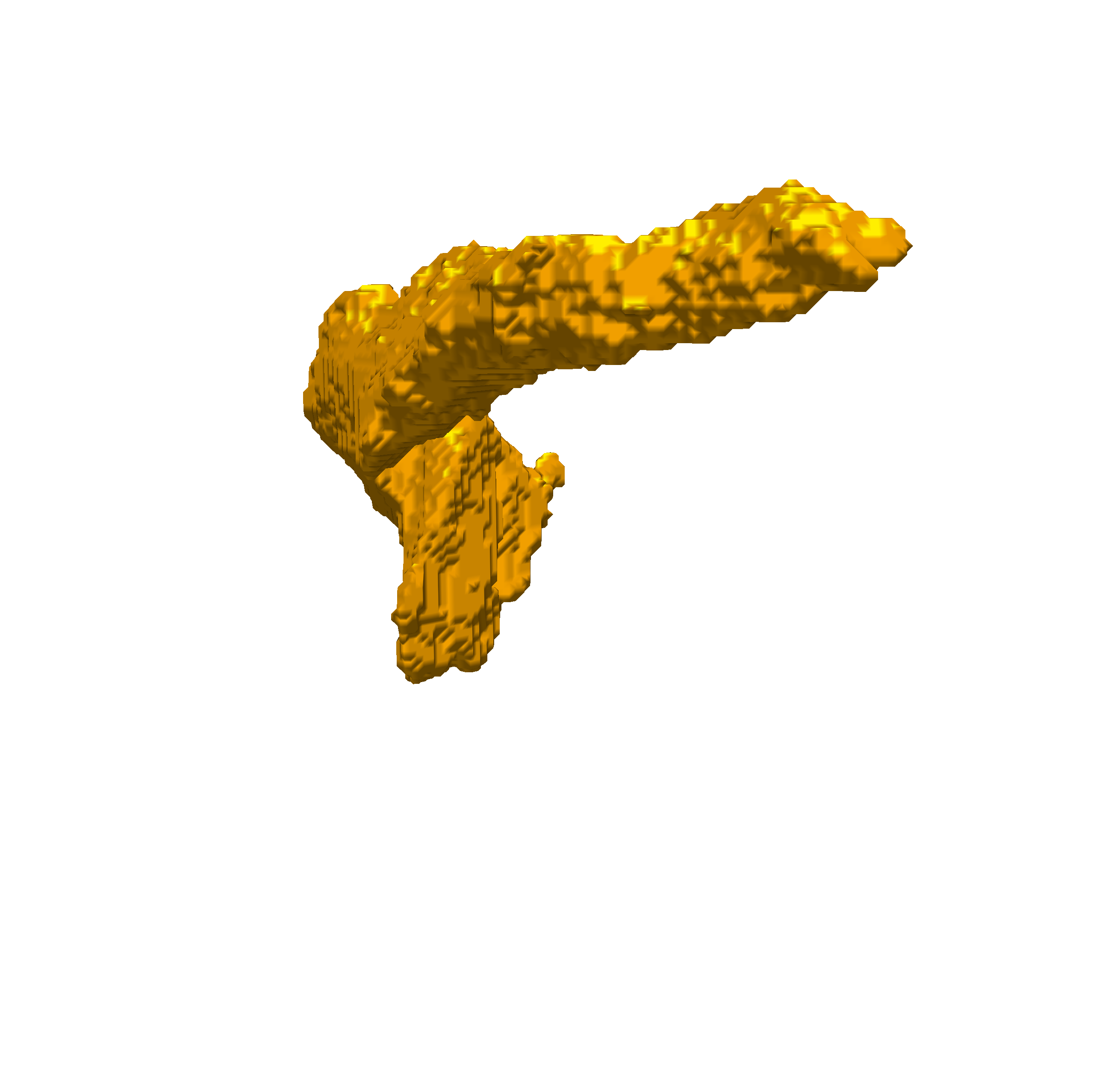}}    	

	\vspace{1em}	
	\caption{Comparison between ground truth, and the prediction results using the different configurations explored in this work (coronal view). The result of setting (d) sum (aug) runs closest to the ground truth (red). No post-processing is applied to the network's output apart from thresholding the probability map at $p(x)>=0.5$.
	\label{fig:comparison}}
\end{figure}

We evaluate our models using a hard split of 100 training and 47 testing patients, and achieve a Dice score performance of 88.3 $\pm$ 4.0 (range [76.4, 94.1])\% (using concatenation layers), and 89.7 $\pm$ 3.8 (range [79.8, 94.8])\% (using summation layers), respectively in testing (see Fig. \ref{fig:training}). To the best of our knowledge, this is the new state-of-the-art performance in pancreas segmentation as can be seen in our comparison to recent literature in Table \ref{tab:comparison}. However, direct comparison to other methods is difficult due to the different datasets and validation schemes employed. A typical test example is shown in Fig. \ref{fig:comparison} for the different model and training configurations explored in this work.
%%%%%%%%%%%%%%%%%%%%%%%%%%%%%%%%%%%%%%%%%%%%%%%%%%%%%%%%%%%%%%%%%%%%%%%%%%%%%%%%%%%%%%%%%%%%%%%%
%%%%%%%%%%%%%%%%%%%%%%%%%%%%%%%%%%%%%%%%%%%%%%%%%%%%%%%%%%%%%%%%%%%%%%%%%%%%%%%%%%%%%%%%%%%%%%%%
%%%%%%%%%%%%%%%%%%%%%%%%%%%%%%%%%%%%%%%%%%%%%%%%%%%%%%%%%%%%%%%%%%%%%%%%%%%%%%%%%%%%%%%%%%%%%%%%
\begin{table}[htb]
\small
%\footnotesize
%\scriptsize
\centering
\caption{ Comparison to other methods. We list other recent pancreas segmentation work performed on the same/similar datasets. A previous method using regression forest (RF) and graph cut (GC)\cite{oda2016regression} on the same dataset, a method utilizing the original Caffe implementation of cascaded 3D U-Nets (trained on a larger dataset\cite{roth2017hierarchical}), and two 2D FCNs methods\cite{roth2017spatial,zhou2016pancreas} utilizing different data, are shown. Validation of other methods was performed using either leave-one-out-validation (LOOV) or cross-validation (CV).}
\label{tab:comparison}
  \vspace{1em}
\begin{tabular}{l|l|l|l|l}
\hline
\rowcolor[gray]{.9}\textbf{Method} & \textbf{Subjects} & \textbf{Approach} & \textbf{Validation} & \textbf{Dice (mean$\pm$std. [min.,max.])\%}\tabularnewline
\hline
\textbf{Proposed (concat)} 										& 100/47 	& RF + 3D FCN 		 & train/test & 88.3 $\pm$ 4.0 [76.4, 94.1]\tabularnewline
\rowcolor[gray]{.9}\textbf{Proposed (sum)} 		& 100/47 	& RF + 3D FCN 		 & train/test & \textbf{89.7 $\pm$ 3.8 [79.8, 94.8]}\tabularnewline
\textbf{3D U-Net (original)}\cite{roth2017hierarchical} 									& 280/150 & 3D FCN (tiling)  & train/test	& 82.2 $\pm$ 10.2 [1.8, 92.2]\tabularnewline
\rowcolor[gray]{.9}\textbf{Oda et al. (2016)}\cite{oda2016regression} & 147 		& RF + GC			 		 & LOOV				& 75.1 $\pm$ 15.4 [18.9, 92.7]\tabularnewline
\textbf{Roth et al. (2017)}\cite{roth2017spatial} 									& 82 			& 2D FCN 					 & 4-fold CV 	& 81.3 $\pm$ 6.3 [50.7, 89.0]\tabularnewline
\rowcolor[gray]{.9}\textbf{Zhou et al. (2016)}\cite{zhou2016pancreas} 									& 82 			& 2D FCN 					 & 4-fold CV 	& 82.4 $\pm$ 5.7 [62.4, 90.9]\tabularnewline
\hline
\end{tabular}
\end{table}
%%%%%%%%%%%%%%%%%%%%%%%%%%%%%%%%%%%%%%%%%%%%%%%%%%%%%%%%%%%%%%%%%%%%%%%%%%%%%%%%%%%%%%%%%%%%%%%%
%%%%%%%%%%%%%%%%%%%%%%%%%%%%%%%%%%%%%%%%%%%%%%%%%%%%%%%%%%%%%%%%%%%%%%%%%%%%%%%%%%%%%%%%%%%%%%%%
%%%%%%%%%%%%%%%%%%%%%%%%%%%%%%%%%%%%%%%%%%%%%%%%%%%%%%%%%%%%%%%%%%%%%%%%%%%%%%%%%%%%%%%%%%%%%%%%
%%%%%%%%%%%%%%%%%%%%%%%%%%%%%%%%%%%%%%%%%%%%%%%%%%%%%%%%%%%%%%%%%%%%%%%%%%%%%%%%%%%%%%%%%%%%%%%%
\section{CONCLUSIONS}
Our models achieve the new state-of-the-art performance in automated pancreas segmentation in CT with close to 90\% average Dice score. This is one of the first studies employing 3D FCNs in a multi-GPU setting for dense volumetric segmentation of the pancreas. 

We showed that deep 3D FCNs can benefit from larger amounts of GPU memory which facilitates mini-batch processing across several training volumes. The utilization of summation layers rather than concatenation layers seems to be beneficial for certain applications such as pancreas segmentation. In the future, the availability of even larger amounts of GPU memory and increased datasets will likely further improve the performance for automated organ segmentation in medical imaging. Our methods could easily be extended to the multi organ segmentation tasks by adjusting the loss function. In the future, some anatomical constraints could be included in order to guarantee topologically correct segmentation results \cite{oktay2017anatomically}.
%%%%%%%%%%%%%%%%%%%%%%%%%%%%%%%%%%%%%%%%%%%%%%%%%%%%%%%%%%%%%%%%%%%%%%%%%%%%%%%%%%%%%%%%%%%%%%%%
%%%%%%%%%%%%%%%%%%%%%%%%%%%%%%%%%%%%%%%%%%%%%%%%%%%%%%%%%%%%%%%%%%%%%%%%%%%%%%%%%%%%%%%%%%%%%%%%
\acknowledgments

This paper was supported by MEXT KAKENHI (26108006, 26560255, 25242047, 17H00867, 15H01116) and the JPSP International Bilateral Collaboration Grant. 
%%%%%%%%%%%%%%%%%%%%%%%%%%%%%%%%%%%%%%%%%%%%%%%%%%%%%%%%%%%%%%%%%%%%%%%%%%%%%%%%%%%%%%%%%%%%%%%%
%%%%%%%%%%%%%%%%%%%%%%%%%%%%%%%%%%%%%%%%%%%%%%%%%%%%%%%%%%%%%%%%%%%%%%%%%%%%%%%%%%%%%%%%%%%%%%%%
\clearpage
\newpage
%\small
\bibliographystyle{spiebib}
\bibliography{spie2018_references}
\end{document}